# "VIRAL" TURING MACHINES, COMPUTATION FROM NOISE AND COMBINATORIAL HIERARCHIES

Theophanes E. Raptis[1,2]

[1]Computational Applications Group, Division of Applied Technologies, National Center for Science and Research "*Demokritos*", Athens, Greece.

[2]Physical Chemistry Lab., National Kapodistrian University of Athens.

E-mail: rtheo@dat.demokritos.gr

**Abstract:** The interactive computation paradigm is reviewed and a particular example is extended to form the stochastic analog of a computational process via a transcription of a minimal Turing Machine into an equivalent asynchronous Cellular Automaton with an exponential waiting times distribution of effective transitions. Furthermore, a special toolbox for analytic derivation of recursive relations of important statistical and other quantities is introduced in the form of an Inductive Combinatorial Hierarchy.

## 1. Introduction

The last three decades, a new computational paradigm known as the interaction computational paradigm (ICP) [1], [2], [3], [4] has slowly arisen out of the birth of the internet at the last two decades of 20[th] century. This has a lot to offer in the case of interacting systems in physics, biology and elsewhere from a purely informational viewpoint wherever a degree of information processing inside individual components in a network is present. Moreover, such informational storage and processing capacity might also prove possible even at a fundamental level of biochemistry [5], [6]. Applications could range from biochemistry to the new field of soft robotics and nano-robotics.

Interactivity is ubiquitous in very many areas of complex systems dynamics. Stephen in [7] as well as Stephen and Dixon in [8] provide strong evidence towards fractal scaling in anticipatory behavior associated with waiting times in the context of organism

– environment interaction. Grigolini also presents in [9] a thorough discussion on the biological emergence issue as a reason for moving beyond mere reductionism based on arguments from renewal processes [10] and anomalous diffusion [11] indicating mechanisms for breakdown of ergodicity. Last year, Hu *et al.* [12] also reported similar effects for single protein molecules in realistic, long *in vivo* times out of their intrinsically fractal energy landscape.

In the case of the ICP, original work by Wegner as well as more recently by Goldin on multi-tape persistent TMs, present a mathematical model of computation in the case of interruptive computation, intermittently forced to receive external inputs and/or provide appropriate outputs on a network or inside an arbitrary noisy environment. Wegner has even proposed that the ICP represents an example of "*Hypercomputation*" or "*Super-Turing Computation*" that moves beyond the confines of the so called "Turing Tarpit" or the associated but as yet formally unproved "Church-Turing Thesis" [13] claiming that in general, ICP contains certain non-algorithmic aspects. This particular issue proved quite thorny and it soon became a matter of fierce debate that strongly polarized the computer science community as can be seen in Davis [14], [15] as well as in the more recent rebuttal in [16]. Recently, Dodig-Crnkovic suggested an alternate epistemological interpretation of information semantics as ICP in [17], associated with info-computationalism [18].

For the aims of the present report we can borrow the simplest original example by Wegner, known as the "Interactive Identity Machine" (IIM) as presented in [1]. The IIM model is a simple reflective mechanism or transducer such that given an indicator function over a set as a Boolean filter, and an iterator construct (*while…do*) it reads an input from some other assumingly intelligent, environmental resource and passes it to its output thus "mimicking" another agent by borrowing its own intelligent responses. While simplistic in its appearance, this trick allows more complex behaviors when put in some network or game with other intelligent agents. The second important aspect of IPC regarding the distinction of an internal discrete time in comparison with an external continuous time flow is also important for the type of stochastic point processes that will be examined here.

One may already notice a certain caveat in the original non-algorithmic argument from a strictly physical viewpoint in that no one has proven as yet the external environment to be really non-algorithmic or the opposite as in the case of so called, "Physical CTT" [19] (Pan-computationalism) identifying natural and algorithmic processes. However, we intend to show with a modification of the simple Wegner's example that under certain assumptions, pure noise of arbitrary statistics can be considered as a superposition of an at least countable infinity of arbitrary computations that can be filtered out.

To this aim, we will introduce in the next section a model of an Interactive "Viral" Turing Machine (IVTM) which is more complex than the original IIM yet simple enough for a toy model. In section 2, we examine a transcription protocol for the complete arithmetization of the dynamics of such machines and in section 3 we examine the resulting dynamics when interacting with a stochastic environment. In section 4, we expand the arithmetization program towards a more complete treatment of generic automata and string rewriting machines by introducing the notion of combinatorial hierarchies. In the last section we also discuss the possible significance of a network of IVTMs interacting both with themselves and the environment in association with the problems of biological emergence and abiogenesis.

## 2. IVTMs as passive dynamical systems

It is well known that the older living organisms from the general class of virii, carrying only genetic material within an external protective capsule, are more or less in an inert condition like a "living-dead" piece of matter until their proximity to a host allows a phase transition due to the host's environmental temperature increase [20]. Then, the closely packed genetic material falls into a more liquid state allowing it to pass into the host's main body through pores using generic diffusion mechanisms.

In a similar sense we can define an abstract machine model from the original TM definition [21] by an appropriate dissection of certain elements requisite in their original definition. Any such TM requires (a) a set of internal control symbols, (b) a set of tape (memory) symbols a subset of which may stand for input and output symbols, (c) a permanent storage medium as a one dimensional memory, originally called the "tape" and (d) a left or right moving "head" upon the tape in order to read or write symbols. The head may also be of sufficient internal complexity to hold the internal control states as a primitive form of "*if…else*" statements. Given an alphabet for (a) and (b), an arbitrary TM is defined by an appropriate Transition Table for all permissible input and output states.

In the standard TM paradigm, an internal mechanism with a source of energy is assumed to guarantee the uninterrupted motion of the reading/writing head. In the IVTM instead, we shall assume a simplified version merging each tape position with additional symbols for a static version with no head making our reduced inert TM unable to perform iterations on its own. It will be only possible to do so when in touch with an external noise source like environmental heat. It is this pairing of an inert TM and a noise source that shall form a complete Interactive Viral TM (IVTM). In order to simplify the presentation, we shall choose as our toy model the recently introduced Wolfram's minimal (2,3) TM [22] which has also been realized recently as an artificial muscle machine [23 ].

To complete the required formal redefinition we shall have to translate every aspect of the original IVTM into a special version of *Asynchronous Cellular Automata* (ACA). We first define the IVTM total transition map via an additional technique which is often called "*Gödelianization*" or "*Arithmetization*" turning the original table into a list of integers by a change of encoding. The terminology originates in the older Goedel encodings [24],[25] introduced for purposes of theoretical examination of logical proofs and first order logic. Practically, we are interested in all possible ways by which a tuple of *n* integers can be mapped 1-1 in *N* as a single integer. One can discern between three main cases

- *Unbounded Tuples – Gödel code*: for any arbitrary set {$n_1$, …, $n_k$} with undefined upper bound the fundamental theorem of arithmetic allows to always finding a unique integer *n* given a list of *k* distinct primes as $n = \pi_1^{n_1}...\pi_k^{n_k}$.
- *Bounded Tuples-maximal element code:* for any set of integers with a given maximal element $n_{max}$ as an upper bound for all sets, the maximal element is chosen as the basis $b = n_{max}$ of a new alphabet such that by the polynomial representation there is always a unique integer $n = n_1 + n_2 b + ... + n_k b^{k-1}$.
- *Bounded Tuples-maximal bit code:* for any set of integers with a maximal element $n_{max}$ of which the binary logarithm is defined as $l(n_{max}) = [\log_2(n_{max})] + 1$ there is always a unique integer such that $n = n_1 + n_2 2^l + ... + n_k 2^{kl-1}$.

It is the last (max. bit) encoding which is the most economical to use for practical purposes as for the arithmetization of TMs. The (2,3)-TM in particular uses a ternary alphabet for tape symbols and a binary alphabet for internal control states plus a motion bit for reading the next memory position via a pointer increase or decrease as shown in Table 1. This allows rewriting each column of the transition table as a set of at most 4-bit integers. For this it is also necessary to introduce an additional bit on top of the first column to restore the symmetry of a 4-bit to 4-bit integer transition. This can be used to represent the previous state of the motion bit which remains indifferent during computation. It should also be noticed that the particular tabulation is not unique and for non ordered tuples of *k* integers one can find at least *k*! ways of assignment to a single integer encoding, yet there is sufficient reason for the particular choice presented here in that it simplifies the expression of a set of evolution equations in the following paragraphs.

Unfolding the first column of Table 1 for the additional bit leads to a complete representation of all possible 16 states forcing also the addition of 4 idle states as fixed points during which no action is taken. The resulting integer sequence is the IVTM's original "genome" under an interpretation protocol that gives different meaning to parts of the binary decoding of each state value. Regarding the dynamics of the integer map given in Table 1, one observes that transitions are disjoint from the unreachable set of the four idle states (fixed points) which are unusable as memory states and stand for "null" or "vacuum" states. This is a peculiarity of the particular minimal implementation of Wolfram's TM which has no "halting" state. The particular subset can be omitted by assuming initial sequences from a hexadecimal alphabet only in the remaining set of 12 usable values. Any trajectory starting from this set is then guaranteed to remain in it. In the general case of arbitrary TMs one can always use the set of idle states as the equivalent of halting states (fixed points) of the dynamics thus representing a stable attractor.

Implementation of the above leads to a major structural difference from the original TM definition in that the actions of the head have now been transplanted into the overall memory structure via a higher alphabet. This causes a certain problem with respect to the correct transfer of control from site to site requiring special treatment. To completely elucidate the type of dynamics behind an IVTM we first need a translation into the language of its ACA equivalent. We observe that in the ACA representation each cell keeps a separate copy of a static head not being capable of any motion. Instead, in order to use the new genomic sequence as the unique transition map ("Rule" in CA parlance) a control signal must be send from each cell to either its left or right neighbor depending on the current cell value taken (*mod* 2). Initialization of a cell array then poses a problem with respect to the values of the control bits for all cells of the static equivalent in such a way that for any array of length $L$ there would be in total $2^L$ possible arrangements, much like a quantum version of the same automaton. In order for this to collapse to a single pointer trajectory, one must take care for transferring also the values of the control bits to the next nearest neighbor by appropriately modifying its previous random value.

For the particular arrangement of Table 1, it is fairly easy to take this last need into account by simply checking whether previous and next cell values belong both to the same half-interval, either $\{2,\ldots,7\}$ or $\{10,\ldots,15\}$ with a 0 or 1 control bit respectively on top, introducing an appropriate indicator function $s(x_t^P, x_t^{P\pm 1}) \in \{+1, 0, -1\}$ where $P$ is a pointer at a particular cell value at time $t$. The auxiliary function $s$ can be used to change the arbitrary value of the control bit of the next cell to that of the previous one whenever there is a transition between half-intervals while it takes a 0 value for both neighbors being on the same half-interval. Such a trajectory can be given analytically for an array of integers as

$$x_{t+1}^{P_t} = T\left(x_t^{P_t} + 8s\left(x_t^{P_t}, x_t^{P_{t-1}}\right)\right) \tag{1a}$$

$$P_{t+1} = P_t + 2\,\mathrm{mod}\left(x_t^{P_t}, 2\right) - 1 \tag{1b}$$

In (1a) we notice that further simplification is possible by taking a global map over all 256 possible input pairs of the abstract functional composition $R(x, y) = T(x + 8s(x, y))$, $x, y \in \{0,\ldots,15\}$. Then we can rewrite (1a) in the more compact form

$$x_{t+1}^{P_t} = R\left(x_t^{P_t}, x_t^{P_{t-1}}\right) \tag{2}$$

In this last version, $R$ can be interpreted as an expanded genome of 256 positions (8-bit input to 4-bit output) where the correction function takes the form

$$s(x_t^{P_t}, x_t^{P_{t-1}}) = (\|x_t^{P_t} - x_t^{P_{t-1}}\| > 7)(2(x_t^{P_t} < x_t^{P_{t-1}}) - 1) \quad (3)$$

One then gets a single 256 symbols sequence to be applied as an operator of the form $R(x + 16y)$. Initial conditions are automatically taken care of via the identity $R(x,x) \cong T(x)$. The unfolded genomic sequence form is shown in Figure 2. In Figure 3, an example of the evolution of such machine in the ACA reformulation is shown for a random initial condition. No boundary conditions have been defined as the original prescription refers to an infinite tape but in physical applications one could consider the case of periodic B. C. (Codes for the examples are available in the account github.com/rtheo/IVTM/) After these formal redefinitions it is explained how this can give rise to a stochastic dynamics for the complete IVTM.

### 3. Stochastic IVTM dynamics and waiting times distributions

To better explain the idea behind the type of interaction envisioned here we should first analyze a simpler example. Let us assume then, an originally deterministic map $f : N \to N$ representing an appropriately constrained real system such that $f$ describes collectively all the allowed possible transitions via an integer encoding. The original system, lacking any internal energy resources would remain inert in the absence of environmental noise, yet at some point in time noise causes a transition from a random initial state $i \to j = f(i)$. We may regard each individual state $i$, $j$ as a set of allowed "conformations" of the underlying construct which could also be a molecular structure including a special type of sensor that acts as a filter which is just a "matching function". Equivalently, one could conceive of a special minimal circuit with a noise harvesting antenna and a memory but with the standard digital processor being replaced by an appropriate filter. Given any noisy environment of arbitrary statistics, the sensor performs a sampling by filtering out an amount of bits at any given time via some threshold device. If a certain sampled value matches the output state in the present conformation $i$ then it sends the device in the new conformation $j$ thus pulling back the predicted output state as the new input state.

In order to give sufficient meaning to the locally causal nature of such a transition with a direct translation of the matching filter to a physical substrate we may borrow the notion of *"affinity"* from similar work in biomacromolecules as already studied in the context of immunochemistry and biochemistry [26], [27]. There the binding affinity is used to express the rate of binding between a ligand and a protein which strongly depends in the geometric structure of the molecules and is often quantitatively expressed via the $L_1$ norm known as Hamming distance given appropriate encodings of the molecular geometry in some symbolic alphabet. In a more primitive context, we may still talk of an affinity of a primitive IVTM device with respect to a given profile of positions and momenta in the total phase space where the IVTM is immersed. Thus, given a countable set of certain preferable local distributions of canonical variables, a given IVTM is sensitive to each one of them for performing a specific transition according to its internal "Rule" genome. If a certain partition of at least an "interesting" subset of the total phase space is possible allowing such a 1-1 correspondence with the defining sequence of an IVTM then it is possible to apply a slightly different variant of the previous deterministic map directly into the ACA formulation of any IVTM as given by (1b) and (2). The crucial difference between a simple deterministic map and an IVTM can then be seen as follows.

Whereas a simple map alters all of the present conformation to a new one, an IVTM handles a single neighborhood of a much larger memory structure at any one time. It is possible to give a precise meaning to this difference using a redefinition of the original evolution equations of the ACA reformulation as follows. Given infinite precision, a generic IVTM system can be further compacted into a unique arbitrarily large integer $N$ as a recursive formula of the form

$$N_{t+1} = N_t + b^{P_t}\left[R(\sigma_{P_t}(N_t), \sigma_{P_{t-1}}(N_t)) - \sigma_{P_t}(N_t)\right] \quad (4)$$

In (2) we use $\sigma_P(n)$ as a symbol extraction operator in any arbitrary base $b$ which can be given analytically as $\sigma(n,P) = \lfloor \mathrm{mod}(n,b)b^{-P} \rfloor$ with the pointer counting from zero position. The above simply replaces each particular value in the polynomial representation. Given that the replacement is analogous to a kind of map differential that can be precomputed from the original map data it can be replaced by

$$N_{t+1} = N_t + b^{P_t}\delta R(\sigma(N_t, P_t) + 16\sigma(N_t, P_{t-1})) \quad (5)$$

where $\delta R(n) = R(n,m) - n, \forall n,m \in \{0,...,15\}$. For this scheme to work it is necessary to provide as an additional constraint that one site is active at each time interval of interaction at least for any given TM specification, yet this is no real restriction as there are equivalent formulations for multi-head and multi-tape machines [28] [29]. Each site, being it active or passive should be capable of a number of possible distinct conformations equal to the highest alphabet value which in practice may be possible by utilizing physical substrates with certain geometric or dynamic group structures.

Such a kind of intermittent dynamics will be characterized by an exponential waiting time distribution between instances of matching which is characteristic for compound Poisson processes and generic Levy point processes. Assuming a 4-bit random input from some noise source appropriately scaled, the matching filter can now be described as

$$Y_t = M(X_t, X_{t'}, X_{t''}) = |R(\lfloor 16X_{t-t'} \rfloor, \lfloor 16X_{t-t''} \rfloor) - X_t| < \varepsilon, t > t' > t'' \quad (6)$$

where $\varepsilon < 2^{-4}$, $X_t$ denotes the underlying continuous random variable and $Y_t$ the new modulated Boolean random variable, prescribing a class of spatially inhomogeneous compound processes. The sequence of $\delta t$ intervals or "waiting times" between transitions is known to obey a probability distribution function

(PDF) analogous to $\exp[\lambda(|\mathbf{r}|)t(m-m_0)]$ with |*r*| denoting the standard Euclidean measure and $\lambda$ a characteristic function for the process which here will depend on *R*, also known as the "Intensity Measure"[30]. In Figures 4a-b, we show two examples of a random sampling under the filter in (6) which show a characteristic shape in a log-log scale. Both under flat noise (a) as well as Brownian noise (b) exhibit a fat tail with an abrupt cutoff. Experimenting with a serial equivalent (program *"itvm.m"*) shows a severe slow down for large sampling times for any other noise source which is not flat.

There are a number of important features to be noticed for such a set of interlinked random processes. First of all, we must make a strict distinction between the ability of any passive system to extract a sequence of computations from noise as a complement to already existing work on the so called, *"noisy computation"* [31], [32], [33], [34]. The latter is dealing with the opposite problem of any such standard sequence of steps being perturbed and/or even prevented from noise to always reach a correct result. In our case though $Y_t$ represents instances of a deterministic process which if faithfully followed would always reach a correct result should such a result exist as a stable attractor. In the last statement we take into account that for this particular class there is no predictability in principle due to the well known uncomputability of the halting problem (*"entscheidungsprobleme"*) [35], [36].

The set of $Y_t$ instances represents a filtered probabilistic set which may or may not end up being stationary depending on both the TM description and the initial random string. To guarantee that such a process always exist one must notice that the waiting time statistics prescribes certain time intervals for all possible strings inside any $\{0,\ldots,b^n\}$ interval which are generally inhomogeneous. The whole process is then characterized by an additional distribution of waiting times per string-to-string path given as a histogram over a real interval $[0, b^n]$ partitioned in $b^n$ subintervals. Additionally, the total histogram is characterized by the ratio of possible fixed points introduced from the arithmetization procedure over the total number of available states. For the (2,3)-TM the characteristic ratio is ¼ while for other TM may vary arbitrarily. Given that the string-time distribution does not diverge for any and all subintervals, then all strings are reachable within finite time from any other. For this it is sufficient that the original $X_t$ PDF does not contain gaps in any subinterval. Extension of the arithmetization paradigm for general functional and combinatorial logic will require the introduction of a special class of combinatorial hierarchies for ordered dictionaries of symbolic strings which is introduced in the last section.

## 4. Global Maps and Inductive Combinatorial Hierarchies

While the previous example sets a generic but simple argument, the inverse problem of classifying all possible machine descriptions from the set of bounded sequences of positive integers and their global maps over increasing length of initial random strings is a much more difficult one. Additionally, the combined study of similar systems under noisy computations diverting them from the correct path either due to weakly constrained implementation, or via unforeseen structural changes, collisions and such, leads to a complicated combinatoric problem. To this aim we introduce a special toolbox of which we will give a general description appropriate for a great variety of problems ranging from CA [37], Ising models [38] up to Designs [39] and general combinatorics while specific details which are beyond the scope of the present report and are postponed for future work.

Combinatorial hierarchies were introduced in an early work by Parker-Rodes [40] and later used by Bastin, Kilmister and Noyes [41], [42] for a different purpose in an effort to link certain combinatoric and number theoretic properties with physics at a fundamental level. A softer version, more appropriate to our purpose is often met in studies of finite grammars and formal languages as a lexical hierarchy of string sets [43]. In what follows we intend to define a simple enough set of Inductive Combinatorial Hierarchies (ICH) of which the power comes from the possibility of defining a set of global maps over successive exponential intervals from which then one can extract by inductive inference simplified laws or algebraic expressions whenever such exist, for the description of ever larger sets.

The fundamental ingredient for any ICH is a set of ordered dictionaries of strings of any given length $n$ in any alphabet $b$. These coincide with the lexicographically ordered combinatoric powersets of any symbolic string which can then be identified with a set of $n \times b^n$ matrices identical with the output of a set of $n$ counters of exponentially increasing period. In the case of the binary alphabet this is also identical with the sampling of a set of oscillators often referred to as the "*Rademacher System*"[44]. A tower of such matrices of increasing length $n$ then will be said to form an ICH in base $b$. We may denote this as $\{L_b(n)\}_{n=1}^{\infty}$ where $L_b(n)$ denotes the dictionary of order $n$ in base $b$, their hierarchical ordering being given by a set of inclusions as $L_b(1) \ldots \subset L_b(n) \subset L_b(n+1) \subset \ldots$ .

Any member of an ICH contains two dual row-wise vs column-wise orders, the first being associated with uniform computability over any $\{0,\ldots,b^n\}$ interval, while the second with individual computability over any string of length $n$. The significance of the distinction will become apparent after the appropriate definitions in the next paragraphs. It is important to distinguish between such combinatoric powersets and set-theoretic or topological powersets os a set $A$ which are always reducible to binary sets as $2^A$. It is always possible to establish a connection between the two by taking $A = \{0,\ldots,b\text{-}1\}$ and an appropriate reduction operator identifying among all $b^n$ strings those with the same set of unique symbols belonging to the same equivalence relation. Then any member of an ICH is reducible to exactly $2^n$ classes, a property common with Closure Spaces [45]. Notably, Griffor in [46] used a similar technique to identify a formal equivalence between string algebras and Clifford algebras.

Assume then any automaton or string rewriting system of which the productions can be given as a set of endomorphisms $\{g_i\}$ from an abstract set of strings $\Sigma$ in any alphabet unto itself. Let then $v \in N$ and let $p(\mathbf{s}) = \sigma_0 + b\sigma_1 + ... + b^{n-1}\sigma_n$ a map from the set of all strings in the b-ary powerset of the alphabet $\sigma \in \{0,...,b-1\}$ to the associated integers via the polynomial representation, the order of the polynomial being defined by the b-ary logarithm $l_b(v) = [\log_b(v)] + 1$. Then, for any string $\mathbf{s} = (\sigma_0, \sigma_1, ..., \sigma_n) \in \Sigma$ there is a mapping $p^{-1}(v) \to \mathbf{s}$. The pair of the $p$ map and its inverse is then called a *b*-ary decoder and encoder respectively. We then define the Complete Arithmetization Program (CAP) as the closing of the square of morphisms by finding a homeomorphism $f_i(v) \cong (p^{-1} \circ g_i \circ p)(v)$ as shown in the diagram.

$$\mathbf{s} \xrightarrow{g_i} \mathbf{s}'$$

$$p^{-1} \uparrow \qquad \downarrow p$$

$$v \xrightarrow{f_i} v'$$

Moreover, given the self-similar nature of any set of counters defining the underlying ICH, we can also define a property of "*subharmonic sensitivity*" if there exists a set of reproducing maps *K* such that

$$f(v) = K\big(f(v - k_j b^{p_i})\big), p_i \in [0,...,n], k_j \in N \qquad (7)$$

In (7), the coefficients $k_j$ are divisors of a finite maximal exponential interval and hence, they must be factors of each alphabet base *b*. Furthermore, if a complete regular sequence of such coefficients and periods $p_i$ exist it may be possible in many cases to extract analytical recursive formulas for the resulting integer sequences corresponding to the requested formula *f(v)* assuming an inductive property holds such that (7) remains invariant for *n+1*. In all such cases a universal, uniformly

computable recursive formula can be found as a vectorized concatenation in the form

$$\{\Sigma_{k+1}\} \leftarrow \{\{\Sigma_k\}, K_1(\{\Sigma_k\}),..., K_{b-1}(\{\Sigma_k\})\} \quad (8)$$

In (8), the action of the reproducing maps is meant to be componentwise to all elements of a set $\Sigma$, for any initial set $\Sigma_0$. One often finds that semi-linear maps like $K_i(x) = a_i x + b_i$ suffice for a variety of different cases like weighted sums of symbols. Details of such function shall be given in forthcoming publications.

One of the simplest examples with non-trivial implications can be given by the so called "*Digit-Sum*" function [47] which increases by one for each exponential interval thus being uniformly computable via the recursion $S_{n+1} \leftarrow \{S_n, S_n + 1\}, S_0 = \{0\}$. An individually computable formula has been found by Trollope-DeLange in [48] via a fractal function. That this is a non-trivial result can be seen with some basic examples involving the entropy and free energy of binary or higher alphabet systems as follows. Given any sample of constant string length over an ICH, all entropic terms can be given as functions of their digit-sums $s_b$ as

$$p_i \log p_i = s_b(v_i)[\log s_b(v_i) - \log(N)] / N \quad (9)$$

Then the total Shannon entropy over any constant string length for any level of the hierarchy can be expressed analytically as

$$\Sigma_{Sh} = \langle s_b(v_i) \log s_b(v_i) \rangle - c \langle s_b(v_i) \rangle, \quad c = \log(N)/N \xrightarrow[N \to \infty]{} 0 \quad (10)$$

The associated Free energy has also been shown to be analogous to the more general Renyi entropies [49] through $F = \Delta T S_\lambda$ where the exponent $\lambda$ is analogous to the "Carnot factor" $T/T_0 = 1 - \Delta T/T$. Then one obtains the equivalent fractal expansion using $s_b$ as

$$F = T \sum_{i=1} s_b(v_i)^\lambda - f(N, \lambda) \quad (11a)$$

$$f(N, \lambda) = \frac{\log N\lambda}{1 - \lambda} = \frac{T_0}{\Delta T} \log N \quad (11b)$$

Expressions (10) and (11) are the CAP equivalents which allow a poweful systematized approach of any similar functions across an ICH.

In more abstract terms, expansion of (8) is formally identical with the individual product terms from a multinomial expansion of a set of $b$ non-commuting functional operators $(K_0 + K_1 + ... + K_{b-1})^k (\Sigma_0)$ where we added the identity map as $K_0$. The same logic can be used for expressing abstract functional programs. If the resulting sequence is also solvable by some generating function than the resulting individually computable formula is necessarily identical to the original automaton and can be used to replace it for all practical purposes. This observation opens the possibility for a new paradigm of *algebraic programming* or equivalently, of *dynamical systems programming*.

The main aim of the CAP scheme is to find ways to eliminate the intermediate path in the square of morphisms as in Figure 5 thus deriving a direct relation in the form of a computable formula $v' = f(v)$ as a representative of any particular class of morphisms avoiding intermediate decoding and encoding. We may call this, the *sequence representation of a powerset of productions*. The importance of resolving the action of an automaton this way stems from the fact that in such cases, an arbitrary production of a given set of morphisms for any automaton could be replaced directly with a dynamical system of lesser complexity although a universal estimator for such a reduction is still lacking. This is practically possible in many cases in an ordered hierarchy of recursively enumerable sets which then allows extracting a single integer sequence for which a simpler recursive formula either uniformly or individually computable over any sequence of exponential intervals can be found. The transcription procedure of a (2,3)-TM to the equivalent ACA model is a restricted primitive example of the CAP proposal while the lack of self-similarity in the final $R$ map is due to the fact that it comes from the composition of two other maps one of which (the original transition table) is not symmetric by definition. The general case of compositions of arithmetized maps and their degree of inheritance of their symmetries is a matter of future research.

Regarding noisy computations, given a recursive scheme for the arithmetized expression of any map over an ICH, one may introduce a distinction between "*holonomic*" and "*non-holonomic*" computations as follows: given a result which is valid and unambiguous inside some model like ordinary arithmetic, the output of any attempt in computing such with any automaton accepting the same input should be the same independently of the "path" or the history of the calculation thus falling to the holonomic class. A computation process which is either ambiguous due to interaction, or deviates due to environmental perturbations may then depend on the actual system path and should be classified as non-holonomic. In any such case, direct use of the corresponding ICH can help in quantifying the degree of deviation or the ambiguity present as well as possible divergent behavior by the use of error intervals surrounding the assumed correct trajectory of the system in question.

5. **Discussion and conclusions**

In the present work we have provided a general framework under which it appears possible to associate highly entropic, noisy environments with generic computational processes embedded in the noise using as a prototype example a particular minimal Turing Machine (TM). The approach being independent of the specific implementation can be applied to any universal TM thus allowing a vast variety of computations being included in such a scheme. A new concept of inductive combinatorial hierarchies has been also introduced as a toolbox for the systematic exploration of such kinds of dynamics via a generic arithmetization procedure.

Future work with possible significance for biochemical structures as well as nano-robotics should expand the previous including also networks of interacting agents as well as possible backreaction of the resulting dynamics to the environment itself which could alter or modulate in the long term the statistics of the noise source itself in the case of bounded spaces also allowing the manifestation of emergent organization as well as some forms of self-organized criticality [50].

Allowing multiple agents acting upon each other intermittently as in the general ICP example, puts an additional problem in that each agent must be able to read anyone from a set of currently active sites in any given conformation which adds certain implementation problems. One way is by adding some kind of marker to separate them from other idle sites which could be possible by some kind of deformable mechanics in which a process similar to biological macromolecules hide parts of their structures leaving active sites as protrusions or using some special form of charging. Actual, physical implementation also poses the problem of a finite memory length which could either restrict the set of computations performable being itself selected by a genetic algorithm and also allowing for some types of polymerization and depolymerization process akin to what is known as "garbage collection", freeing used memory. Surprisingly, a similar process is reported as dynamic instability of the so called microtubular structures in eukaryotic cells cytoskeletal structures [51], [52]. Notably, a physico-chemical equivalent of Church's λ-Calculus has also been introduced by Buliga as well as Buliga and Kaufman in a recent series of publications [53], [54], [55] which is appealing for testing more general computational processes in the light of the combinatorial hierarchy concepts introduced.

| (0, 1)   | *000 | (L/R)*** | ***  | *000 | (0, 1) |
|----------|------|----------|------|------|--------|
| (2, 3)   | *100 | (L/R)A0  | RB1  | 1011 | 13     |
| (4, 5)   | *010 | (L/R)B0  | LC0  | 0110 | 6      |
| (6, 7)   | *110 | (L/R)C0  | LB0  | 0010 | 4      |
| (8, 9)   | *001 | (L/R)*** | ***  | *001 | (8, 9) |
| (10, 11) | *101 | (L/R)A1  | LC0  | 0110 | 6      |
| (12, 13) | *011 | (L/R)B1  | RC1  | 1111 | 15     |
| (14, 15) | *111 | (L/R)C1  | RA0  | 1100 | 3      |

**Table 1**. Arithmetization of the Transition Table for the Wolframm (2,3) minimal UTM. Asterisks have been put in the indifferent states of the total of $2^4$ lexicographically ordered input states while the motion bit has been added on top.

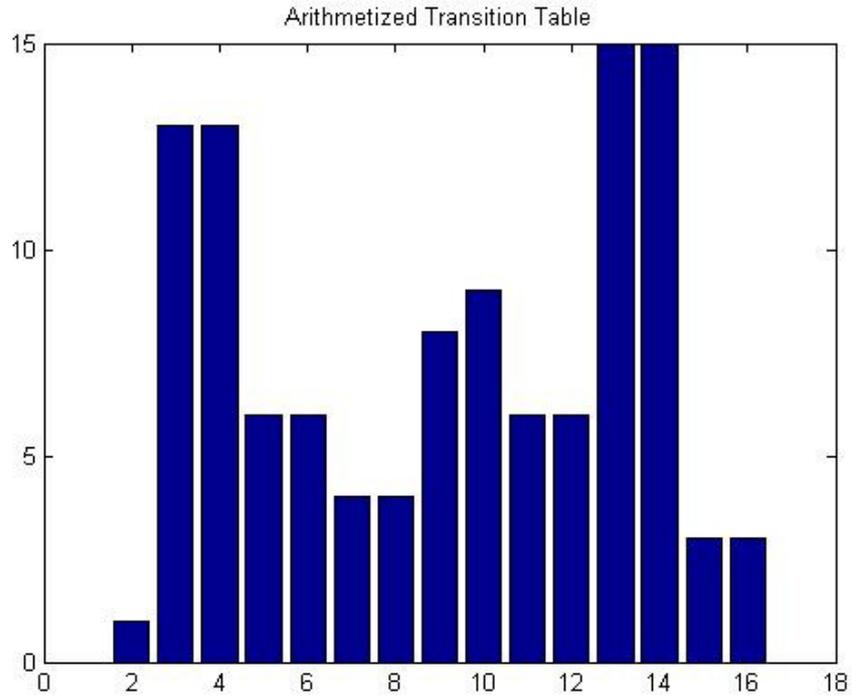

**Fig. 1**. Graph of the resulting sequence representing the total arithmetized transition map of Table 1.

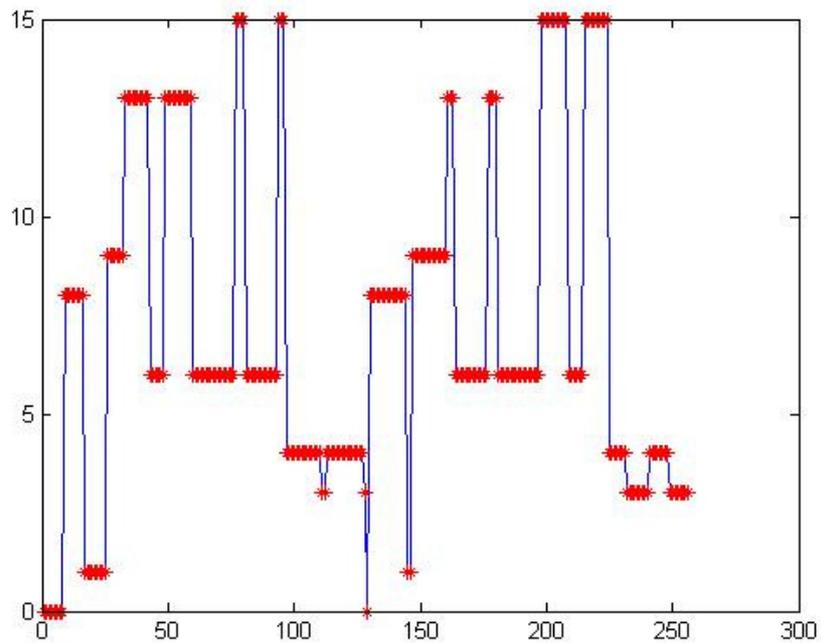

**Fig. 2**. Graph of the expanded "Rule" sequence for the ACA representation of the (2,3)-TM.

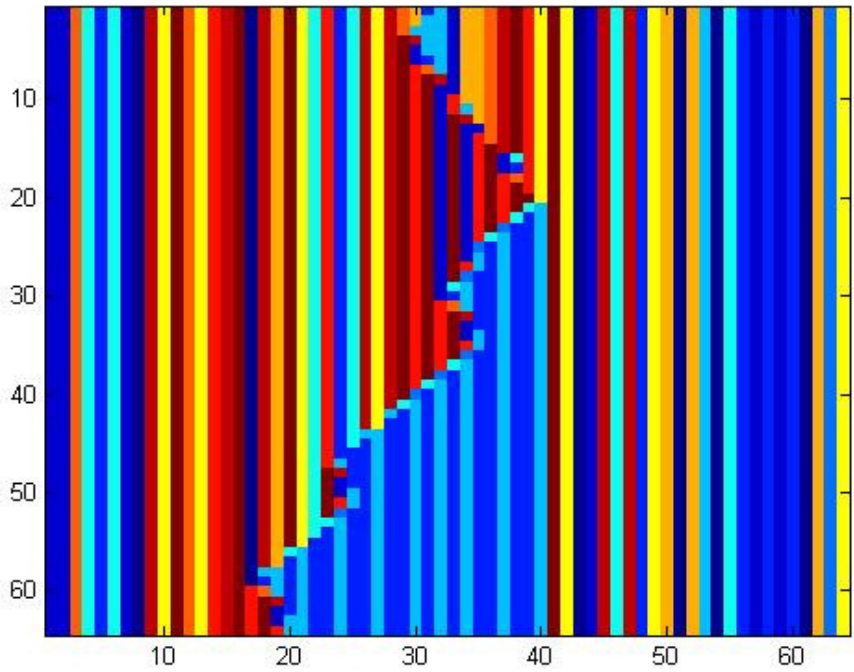

**Fig. 3**. Evolution of a random initial condition for the Asynchronous CA redefinition of the (2,3)-TM.

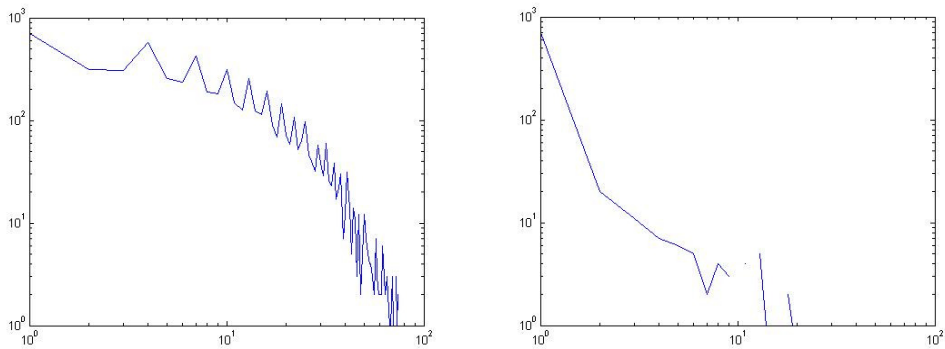

**Fig. 4**. (a) Log-Log plot of histogram for a sampling under the matching filter of section 3 with a flat random generator, (b) the same for a modulated Brownian noise generator.